\documentclass[final]{cvpr}
\usepackage{bibunits}
\defaultbibliography{cvpr}
\defaultbibliographystyle{ieee_fullname}

\usepackage[toc,page,header]{appendix}
\usepackage{minitoc}

\usepackage{times}
\usepackage{epsfig}
\usepackage{graphicx}
\usepackage{amsmath}
\usepackage{amssymb}
\usepackage{nopageno}
\usepackage{multibib}

\usepackage{mathtools}
\usepackage{booktabs}
\usepackage{array}
\usepackage{multirow}
\usepackage{caption}
\usepackage{arydshln}
\usepackage{blindtext}

\makeatletter
\newcommand{\printfnsymbol}[1]{%
  \textsuperscript{\@fnsymbol{#1}}%
}
\makeatother

\usepackage[pagebackref=true,breaklinks=true,colorlinks,bookmarks=false]{hyperref}

\def\cvprPaperID{6778}
\def\confYear{CVPR 2021}

\begin{document}
\begin{bibunit}
\doparttoc 
\faketableofcontents 
\title{PANDA: Adapting Pretrained Features for Anomaly Detection and Segmentation}

\author{Tal Reiss \thanks{ Equal contribution} , Niv Cohen \printfnsymbol{1}, Liron Bergman \& Yedid Hoshen  \\
   School of Computer Science and Engineering \\
  The Hebrew University of Jerusalem, Israel \\
}

\maketitle

\begin{abstract}
Anomaly detection methods require high-quality features. In recent years, the anomaly detection community has attempted to obtain better features using advances in deep self-supervised feature learning. Surprisingly, a very promising direction, using pretrained deep features, has been mostly overlooked. In this paper, we first empirically establish the perhaps expected, but unreported result, that combining pretrained features with simple anomaly detection and segmentation methods convincingly outperforms, much more complex, state-of-the-art methods. 

In order to obtain further performance gains in anomaly detection, we adapt pretrained features to the target distribution. Although transfer learning methods are well established in multi-class classification problems, the one-class classification (OCC) setting is not as well explored. It turns out that naive adaptation methods, which typically work well in supervised learning, often result in catastrophic collapse (feature deterioration) and reduce performance in OCC settings. A popular OCC method, DeepSVDD, advocates using specialized architectures, but this limits the adaptation performance gain. We propose two methods for combating collapse: i) a variant of early stopping that dynamically learns the stopping iteration ii) elastic regularization inspired by continual learning. Our method, PANDA, outperforms the state-of-the-art in the OCC, outlier exposure and anomaly segmentation settings by large margins\footnote{The code is available at github.com/talreiss/PANDA}.

\end{abstract}

\section{Introduction}
\label{sec:intro}

Detecting anomalous patterns in data is of key importance in science and industry. In the computational anomaly detection task, the learner observes a set of training examples. The learner is then tasked to classify novel test samples as normal or anomalous. There are multiple anomaly detection settings investigated in the literature, corresponding to different training conditions. In this work, we deal with three settings: i) anomaly detection - when only normal images are used for training  ii) anomaly segmentation - detecting all the pixels that contain anomalies, given normal images as input. iii) Outlier Exposure (OE) - where an external dataset simulating the anomalies is available.

In recent years, deep learning methods have been introduced for anomaly detection, typically extending classical methods with deep neural networks. Different auxiliary tasks (e.g. autoencoders or rotation classification) are used to learn representations of the data, while a great variety of anomaly criteria are then used to determine if a given sample is normal or anomalous. An important issue for current methods is the reliance on limited normal training data for representation learning, which limits the quality of learned representations. Nearly all state-of-the-art anomaly detection methods rely on self-supervised feature learning - i.e. using the limited normal training data for learning strong features. The motivation for this is twofold: i) the fear that features trained on auxiliary domains will not generalize well to the target domain. ii) the curiosity to investigate the top performance achievable without ever looking at any external dataset (we do not address this question here).

In other parts of computer vision, features pretrained on external datasets are often used to improve performance on tasks trained on new domains - and our reasonable hypothesis is that this should also be the case for image anomaly detection and segmentation. We present very simple baselines that use pretrained features trained on a large external data and K-nearest neighbor (kNN) retrieval to significantly outperform all previous methods on anomaly detection and segmentation, even on images of distant target domains.


We then tackle the technical challenge of obtaining stronger performance by further adaptation to the normal training data. Although feature adaptation has been extensively researched in the multi-class classification setting, limited work was done in the OCC setting. Unfortunately, it turns out that feature adaptation for anomaly detection often suffers from \textit{catastrophic collapse} - a form of deterioration of the pretrained features, where all (including anomalous) samples, are mapped to the same point. DeepSVDD \cite{ruff2018deep} proposed to overcome collapse by removing biases from the model architecture, but this restricts network expressivity and limits the pretrained models that can be borrowed off-the-shelf. Perera and Patel \cite{perera2019learning} proposed to jointly train OCC with the original task which has several limitations and achieves only limited adaptation success. 

Our first finding is that simple training with constant-duration early stopping (with no bells-and-whistles) already achieves top performance. To remove the dependence on the number of epochs for early stopping, we propose two techniques to overcome catastrophic collapse: i) an adaptive early stopping method that selects the stopping iteration per-sample, using a novel generalization criterion - this technique is designed to overcome a special problem of OCC, namely that there are no anomalies in the validation set ii) elastic regularization, motivated by continual learning, that postpones the collapse. Thorough experiments demonstrate that we outperform the state-of-the-art by a wide margin (ROCAUC): e.g. CIFAR10 results: 96.2\% vs. 90.1\% without outlier exposure and 98.9\% vs. 95.6\% with outlier exposure. We also achieve $96.0\%$ vs. $89.0\%$ on anomaly segmentation on MVTec.



We present insightful critical analyses: i) We show that pretrained features strictly dominate current self-supervised RotNet-based feature learning methods. We  discuss the relative merits of each paradigm and conclude that for most practical purposes, using pretrained features is preferable.  ii)  We analyse the results of the popular DeepSVDD method and discover that its feature adaptation, which is designed to prevent collapse, does not improve over simple data whitening. 

\textbf{Contributions:} To summarize our main contributions in this paper:
\begin{itemize}
    \item Demonstrating that a simple baseline outperforms all current methods in image anomaly detection and segmentation - extensive analysis shows the generality of the result.
    \item Identifying that popular SOTA methods do not outperform linear whitening in OCC feature adaptation.
    \item Proposing several effective solutions for feature adaptation for OCC.
    \item Extensive evaluation, obtaining results that significantly improve over the current state-of-the-art. 
\end{itemize}


\subsection{Related Work}
\label{subsec:related}

\textit{Classical anomaly detection:} The main categories of classical anomaly detection methods are: i) reconstruction-based: compressing the training data using a bottleneck, and using a reconstruction loss as an anomaly criterion (e.g. \cite{candes2011robust, jolliffe2011principal}, K nearest neighbors \cite{eskin2002geometric} and K-means \cite{hartigan1979algorithm}), ii) probabilistic: modeling the probability density function and labeling unlikely sampled as anomalous (e.g. Ensembles of Gaussian Mixture Models \cite{glodek2013ensemble}, kernel density estimate \cite{latecki2007outlier}) iii) one-class classification (OCC): finding a separating manifold between normal data and the rest of input space (e.g. One-class SVM \cite{scholkopf2000support}).

\textit{Deep learning methods:} The introduction of deep learning has affected image anomaly detection in two ways: extension of classical methods with deep representations and novel self-supervised deep methods{\cite{pang2021deep}\cite{ruff2021unifying}}. Reconstruction-based methods have been enhanced by learning deep autoencoder-based bottlenecks \cite{dgroup} which can provide better models of image data. Deep methods extended classical methods by creating a better representations of the data for parametric assumptions about probabilities, a combination of reconstruction and probabilistic methods (such as DAGMM \cite{zong2018deep}), or in a combination with OCC methods \cite{ruff2018deep}. Novel deep methods have also been proposed for anomaly detection including GAN-based methods \cite{zong2018deep}. Another set of novel deep methods use auxiliary self-supervised learning for anomaly detection. The seminal work by \cite{golan2018deep} was later extended by \cite{hendrycks2019using} and \cite{bergman2020classification}. 

\textit{Transferring pretrained representations:} Learning deep features requires extensive datasets, preferably with labels. An attractive property of deep neural networks, is that representations learned on very extensive datasets, can be transferred to data-poor tasks. Specifically deep neural representations trained on the ImageNet dataset have been shown by \cite{huh2016makes} to significantly boost performance on other datasets that are only vaguely related to some of the ImageNet classes. This can be performed with and without finetuning. Although much recent progress has been performed on self-supervised feature learning \cite{gidaris2018unsupervised, chen2020simple}, such methods are typically outperformed by transferred pretrained features. Transferring ImageNet pretrained features have been proposed for out-of-distribution detection  by \cite{hendrycks2019pretraining} and video anomaly detection \cite{smeureanu2017deep} \cite{georgescu2021anomaly}. Similar pretraining has been proposed for one-class classification has been proposed by \cite{perera2019learning}, however they require joint optimization with the original task. Rippel et. al. \cite{rippel2020modeling} follow an early version of this paper and report results with pretrained features on MVTec using the Mahalanobis distance. A concurrent work adapts pretrained features for the Outlier Exposure setting \cite{deecke2020deep} and achieves similar results to our method (see Sec.\ref{sec:method_panda}).

\textit{Anomaly segmentation methods:}
Segmenting the image pixels that contain anomalies has attracted far less research attention than image-level anomaly detection. Several previous anomaly segmentation works used pretrained features, but they have not convincingly outperformed top self-supervised methods. Napoletano et al. \cite{napoletano2018anomaly} extracted deep features from small overlapping patches, and used a K-means based classifier over dimensionality reduced features. Bergmann et al. \cite{bergmann2019mvtec} evaluated both a ADGAN and autoencoder approaches on MVTec dataset \cite{bergmann2019mvtec} finding complementary strengths. More recently, Venkataramanan et al. \cite{venkataramanan2020attention} used an attention-guided VAE approach combining multiple methods (GAN loss \cite{goodfellow2014generative}, GRADCAM \cite{selvaraju2017grad}). Bergmann et al. \cite{bergmann2020uninformed} used a student-teacher based autoencoder approach employing pretrained ImageNet deep features. Our simple baseline, SPADE, significantly outperforms the previously mentioned approaches.


\section{A General Framework and Simple Baselines for Anomaly Detection and Segmentation}

\subsection{A Three-stage Framework}

We present our general framework in which we examine several adaptation-based anomaly detection methods, including our method. Let us assume that we are given a set ${\cal{D}}_{train}$ of normal training samples:  $x_1,x_2,\ldots,x_N$. The framework consists of three steps:

\textbf{Initial feature extractor:} 
An initial feature extractor $\psi_0$ can be obtained by pretraining on an auxiliary task with loss function $L_{pretrain}$. The auxiliary task can be either pretraining on an external dataset (e.g. ImageNet) or by self-supervised learning (auto-encoding, rotation or jigsaw prediction). In the former case, the pretrained extractor can be obtained off-the-shelf. The choice of auxiliary tasks is analyzed in Sec.~\ref{sec:DetailedResults}.

\textbf{Feature adaptation:} Features trained on auxiliary tasks or datasets may require adaptation before being used for anomaly scoring on the target data. This is seen as a finetuning stage of the features on the target training data. We denote the feature extractor after adaptation  $\psi$. 

\textbf{Anomaly scoring:} Having adapted the features for anomaly detection, we extract the features $\psi(x_1),\psi(x_2),\ldots,\psi(x_N)$ of the training set samples. We then proceed to learn a scoring function, which describes how anomalous a sample is. Typically, the scoring function seeks to measure the density of normal data around the test sample $\psi(x)$ (either by direct estimation or via some auxiliary task) and assign a high anomaly score to low density regions. 

We report very simple-to-implement but highly effective baselines for anomaly detection and segmentation, based on our framework. In both baselines we skip the adaptation stage. Implementation details of both methods, including the K-Means-based speedup, can be found in the appendix (App.\ref{app:implementation}).

\subsection{Simple Baseline for Anomaly Detection}

In the anomaly detection baseline "Deep Nearest Neighbors" (\textbf{DN2}), the feature extractor is a large ResNet pretrained the ImageNet dataset. The features are taken from the final pooling layer (before the linear classifier). For each test image we use the average distance from the features of the kNN normal images as the anomaly score.

\subsection{Simple Baseline for Anomaly Segmentation}

In the anomaly segmentation baseline, "Semantic Pyramid Anomaly Detection" (\textbf{SPADE}), we use an ImageNet-pretrained ResNet to extract per-pixel features for all images. As both low-level high-resolution features, and semantic low-resolution features are important for determining if a pixel is anomalous, we extract features from multiple layers of the deep neural network.  Specifically, we extract spatial feature maps from the first three blocks of a ResNet network, and associate each pixel with its corresponding descriptors from the maps. To get a single descriptor, we concatenate the descriptors from the three layers. Finally, we score the pixel by its kNN distance from the feature descriptors of the pixels of all training images. In both baselines we skip the adaptation stage. Implementation details of both methods, including the K-Means-based speedup, can be found in the Supplementary Material (SM).

\begin{figure}[t]
\begin{center}
    \begin{tabular}{cc}
   \includegraphics[scale=0.33]{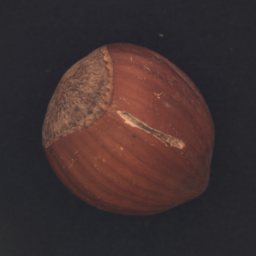} &
   \includegraphics[scale=0.33]{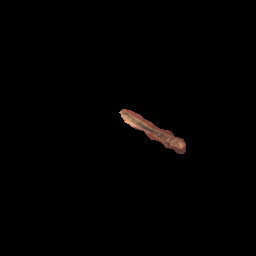} \\

    \end{tabular}
    \end{center}
    \caption{(left) An anomalous image (right) The predicted anomalous image pixels. }
    \label{fig:hazelnut}
\end{figure}

\section{Feature Adaptation for Anomaly Detection}
\label{Framework}

Although our two simple-to-implement baselines, DN2 and SPADE, achieve very strong results, we ask if feature adaptation can improve them further. We first review two existing methods for feature adaption for anomaly detection, and proceed to propose our method, PANDA, which significantly improves over them.   

\subsection{Background: Existing Feature-Adaptation Methods}

\textbf{DeepSVDD:} Ruff et al. \cite{ruff2018deep} suggest to first train an autoencoder on the normal-only train images. The encoder is then used as the initial feature extractor $\psi_0$. As the features of the encoder are not specifically adapted to anomaly detection, DeepSVDD adapts $\psi$ on the training data. The adaptation takes place by minimizing the compactness loss:
\begin{equation}
\label{eq:L_compact}
    L_{compact} = \sum_{x \in {\cal{D}}_{train}} \|\psi(x) - c\|^2
\end{equation}
Where $c$ is a constant vector, typically the average of $\psi_0(x)$ on the training set. However, the authors were concerned of the trivial solution $\psi = c$, and suggested architectural restrictions to mitigate it, most importantly removing the biases from all layers. We empirically show that the effect of adaptation of the features in DeepSVDD does not outperform simple feature whitening (see Sec.~\ref{adapt_anl}).      

\textbf{Joint optimization (JO):} Perera et al. \cite{perera2019learning} proposed to use a deep feature extractor trained for object classification on the ImageNet dataset. Due to fear of "learning a trivial solution in the absence of a penalty for miss-classification", the method does not adapt by finetuning on the compactness loss only. Instead, they  relaxed the task setting, by assuming that a number ($\sim50k$) of labelled original ImageNet images, ${\cal{D}}_{pretrain}$, are still available at adaptation time. They proposed to train the features $\psi$ under the compactness loss jointly with the original ImageNet classification linear layer $W$ and its classification loss, here the CE loss with the true label $\ell_{pretrain}(p,y) = -\log(p_{y})$, and SMax indicates Softmax:
\begin{multline}
\label{eq:pandp}
L_{Joint} =  \sum_{(x,y) \in {\cal{D}}_{pretrain}} \ell_{pretrain}(SMax(W \psi(x)), y) \\ + \alpha  \cdot \sum_{x \in {\cal{D}}_{train}} \|\psi(x) - c\|^2
\end{multline}

Where $W$ is the final linear classification layer and $\alpha$ is a hyper-parameter weighting the two losses.  We note that the method has two main weaknesses: i) it requires retaining a significant number of the original training images which can be storage intensive ii) jointly training the two tasks may reduce the anomaly detection task accuracy, which is the only task of interest in this context.

\begin{figure}[t]

  \centering
 
   \includegraphics[scale=0.25]{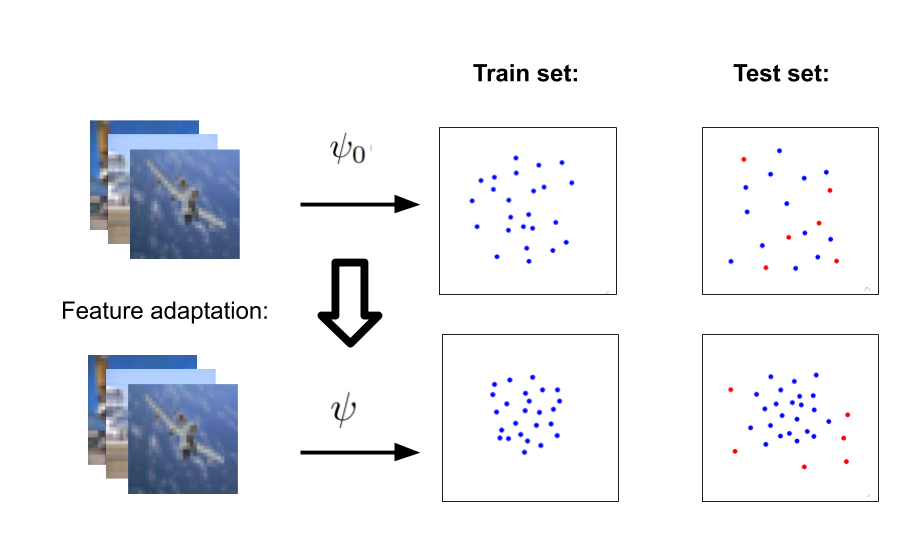}
    \label{fig:diagram}
    \caption{An illustration of our feature adaptation procedure, the pretrained feature extractor $\psi_0$ is adapted to make the normal features (blue) more compact resulting in feature extractor $\psi$. After adaptation, anomalous test features (red)  lie in a less dense region of the feature space.}
    \vspace{-1.25em}
\end{figure}

\subsection{PANDA: pretrained Anomaly Detection Adaptation}
\label{sec:method_panda}
We present PANDA, a new method for anomaly detection in images. Similarly to SVDD and Joint Optimization, we also use the compactness loss (Eq.~\ref{eq:L_compact}) to adapt the general pretrained features to the task of anomaly detection on the target distribution. Instead of constraining the architecture or introducing external data into the adaptation procedure we tackle catastrophic collapse directly. The main challenge is that the optimal solution of the compactness loss can result in "collapse", where all possible input values are mapped to the same point ($\psi(x) =c, \ \  \forall x $). Learning such features will not be useful for anomaly detection, as both normal and anomalous images will be mapped to the same output, preventing separability. The issue is broader than the trivial "collapsed" solution after full convergence, but rather the more general issue of feature deterioration, where the original good properties of the pretrained features are lost. Even a non-trivial solution might 
lose some of the discriminative properties of the original features which are none-the-less important for anomaly detection.

To avoid this collapse, we suggest three options: (i) finetuning the pretrained extractor with compactness loss (Eq.\ref{eq:L_compact}) and stopping after a constant number of iterations (ii) a novel method for determining early stopping per-sample (iii) when collapse happens prematurely, before any significant adaptation happens, we suggest mitigating it using a Continual Learning-inspired adaptive regularization.

\begin{figure*}

  \centering

   \includegraphics[scale=0.165]{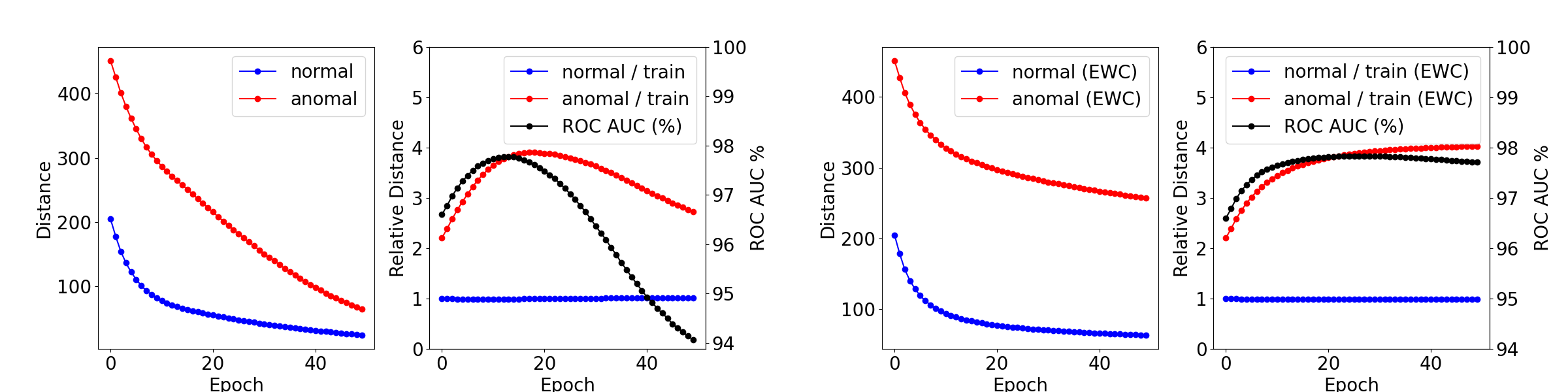} 
    \caption{CIFAR100 Class 17 (left to right): (1) -  During training all samples approach the center of train set features (2) -  When normalized by the train average distance $s_t$, the normal samples stay dense, while the anomalous ones initially move further away and then "collapse". The ROC AUC performance behaves similarly to the anomalous samples' normalized distance. (3),(4) - when training with EWC the collapse is mitigated. }
      \label{time_plots}

\end{figure*}

\textit{Simple early stopping (PANDA-Early)}: An embarrassingly simple but effective solution for controlling the collapse of the original features is to stop training after a constant number of iterations (e.g. $15$ epochs on CIFAR10). Inversely scaling the number of epochs by dataset size works for most examined datasets   (Sec.~\ref{sec:DetailedResults}). 

\textit{Sample-wise early stopping (PANDA-SES)}: A weakness of the simple early-stopping approach, is the reliance on a hyper-parameter that may not generalize to new datasets. Although the optimal stopping epoch can be determined with a validation set containing anomalies, it is not available in our setting. We thus propose "samplewise early stopping" (SES) as an unsupervised way of determining the stopping epoch from a single sample. The intuition for the method can be obtained from Fig.~\ref{time_plots}. We can see that anomaly detection accuracy is correlated to having a large ratio between the distance of the anomalous samples to the center, and the distance between the normal samples and the center. We thus propose to save checkpoints of our network at fixed intervals (every 5 epochs) during the training process - corresponding to different early stopping iterations 
($\psi_1,\psi_2,\ldots,\psi_T$), for each network $\psi_t$ we compute the average distance on the training set images $s_t$. During inference, we score a target image $x$ using each model $s_t^{target} = \|\psi_t(x) - c\|^2$, and normalize the score by the training average score by $s_t$. We set the maximal ratio, as the anomaly score of this sample, as this roughly estimates the model that achieves the best separation between normal and such anomalous samples.


\textit{Continual Learning (PANDA-EWC):} We propose a new solution for overcoming premature feature collapse that draws inspiration from the field of continual learning. The task of continual learning tackles learning new tasks without forgetting the previously learned ones. We note however that our task is not identical to standard continual learning as: i) we deal with the one-class classification setting whereas continual-learning typically deals with multi-class classification ii) we aim to avoid forgetting the expressivity of the features but do not particularly care if the actual classification performance on the old task is degraded. A simple solution for preventing feature collapse is regularization of the change in value of the weights of the feature extractor.  However, this solution is lacking as some weights influence the features more than others.

Following ideas from continual learning, we use elastic weight consolidation (EWC) \cite{kirkpatrick2017overcoming}. Using a number of mini-batches (we use $100$) to pretrain on the auxiliary task. We compute the diagonal of the Fisher information matrix $F$ for all weight parameters of the network. Note that this only needs to happen once at the end of the pretraining stage. The value of the Fisher matrix for diagonal element $\theta$ is given by:
\begin{equation}
    \label{eq:fisher}
    F_{\theta} = \mathbb{E}_{(x, y) \in {\cal{D}}_{pretrain}}\left[\left(\frac{\partial}{\partial \theta} L_{pretrain}(x, y)\right)^2\right]
\end{equation}
We follow \cite{kirkpatrick2017overcoming} in using the diagonal of the Fisher information matrix $F_{\theta_i}$, to weight the squared distance of the change of each network pretrained weight $\theta_i \in \psi_0$ and finetuned weight $\theta_i^{*} \in \psi$. This can be seen as a measure of the loss landscape curvature as function of the weights - larger values imply high curvature, inelastic weights.

We use this regularization in combination with the compactness loss weighted by the factor $\lambda$, which is a hyperparameter of the method (we use $\lambda=10^{4}$):
\begin{equation}
    \label{eq:panda}
    L_{\theta} = L_{compact}(\theta^*) + \frac{\lambda}{2} \cdot \sum_{i}F_{\theta_i}(\theta_i - \theta^*_{i})^2
\end{equation}
The network $\psi$ is initialized with the parameters of the pretrained extractor $\psi_0$ and trained with SGD.


\textbf{Anomaly scoring:}
As in classical anomaly detection, scoring can be done by density estimation. Unless mentioned otherwise, we use kNN for scoring. We also evaluate faster methods and get similar results (see Sec.~\ref{sec:scoring}).
 
\textbf{Outlier Exposure:} An extension of the typical image anomaly detection task \cite{hendrycks2018deep}, assumes the existence of an auxiliary dataset of images ${\cal{D}}_{OE}$, which are more similar to the anomalies than normal data. In case such information is available, we can treat it as a standard classification task and simply train a linear classification layer $w$ together with the features $\psi$ under a logistic regression loss (Eq. \ref{eq:oe}). As before, $\psi$ is initialized with the weights from $\psi_0$. After training $\psi$ and $w$, we use $w \cdot \psi(x)$ as the anomaly score. Results and critical analysis of this setting are presented in Sec.~\ref{sec:DetailedResults}.
\begin{equation}
    \label{eq:oe}
    L_{OE} = \sum_{x \in {\cal{D}}_{train}} log(1 - \sigma(w \cdot \psi(x))) + \sum_{x \in {\cal{D}}_{OE}} log(\sigma(w \cdot \psi(x)))
\end{equation}

\begin{figure*}[t]

  \centering

    \begin{tabular}{cccc}
   \includegraphics[scale=2.4]{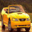} &
    \includegraphics[scale=0.23]{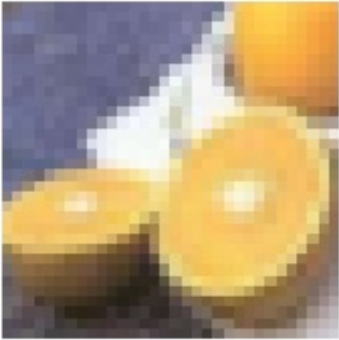} & 
    \includegraphics[scale=0.175]{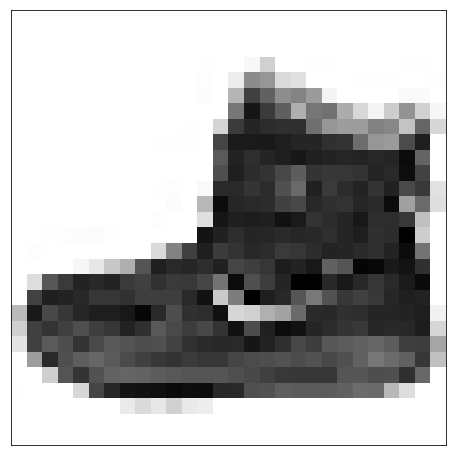} & 
    \includegraphics[scale=0.275]{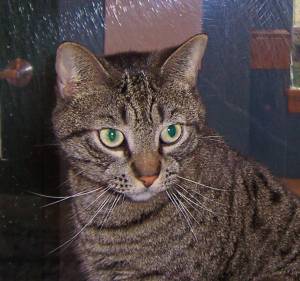} 
    \\
    \includegraphics[scale=0.08]{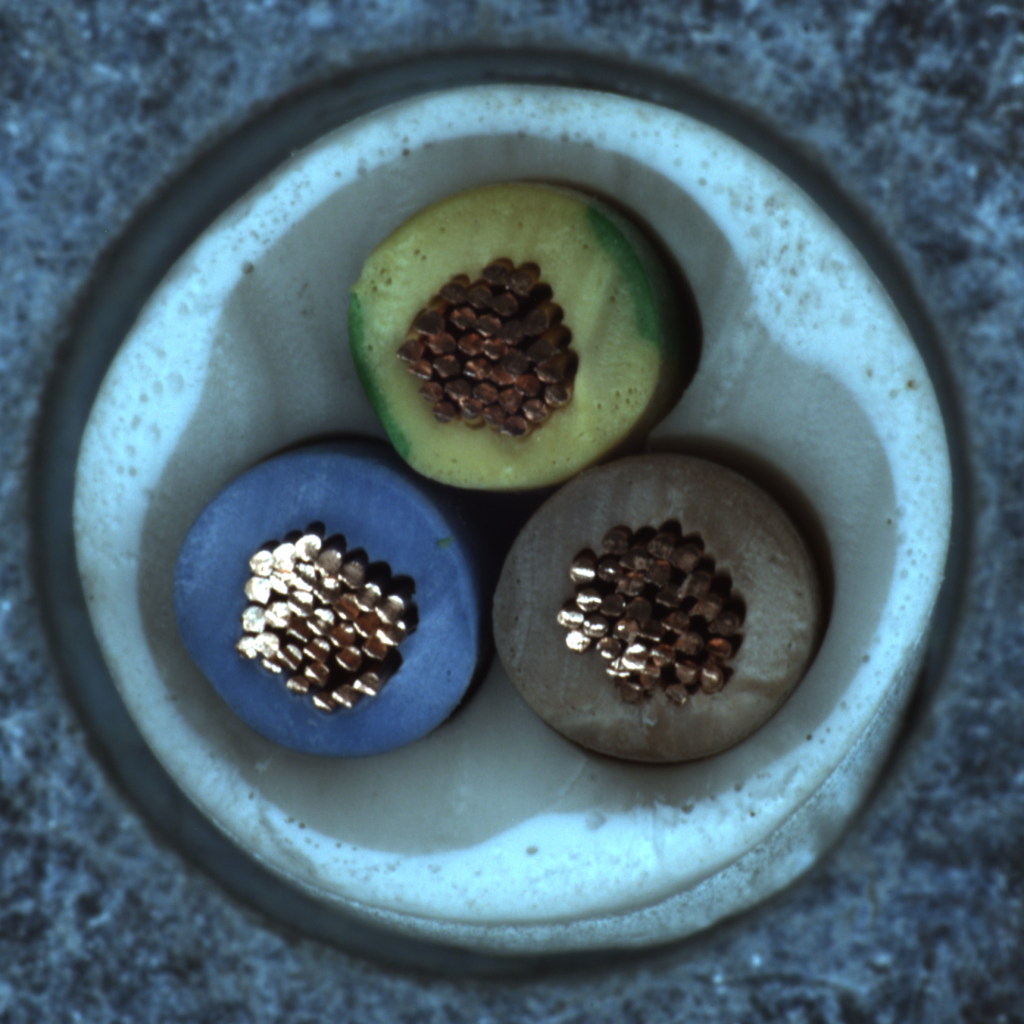} & 
   \includegraphics[scale=0.16]{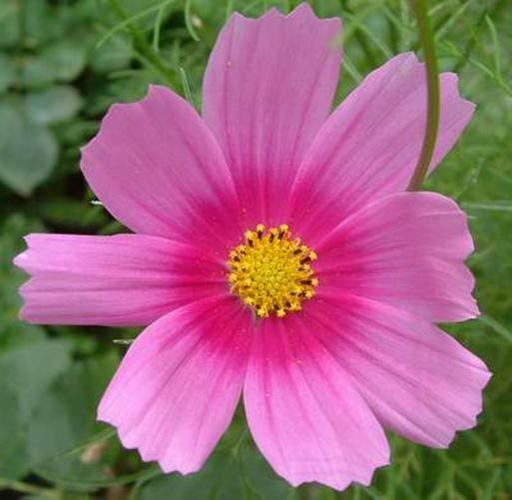} & 
   \includegraphics[scale=0.30]{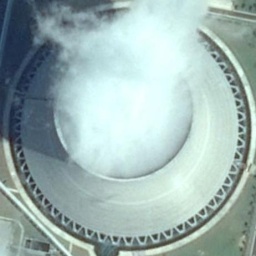} & \includegraphics[scale=0.68]{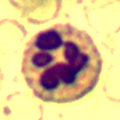}

    \end{tabular}
    \caption{Representative images of the different datasets, from the left clockwise: CIFAR10, CIFAR100, Fashion MNIST, DogsVsCats, MVTec, Oxford Flowers, DIOR and WBC. Following standard protocol, in all datasets (except MVTec), normal data are one class (e.g. cat in CIFAR10) while anomalies are all other test data from the same dataset (e.g. dog, car in CIFAR10). MVTec contains class-specific anomalies (e.g. for normal class - Wire, anomalies include bent wires)}
      \label{fig:dataset_images}
\vspace{-1.0em}
\end{figure*}

\section{Image Anomaly Detection}
\label{sec:Image}

\subsection{Experiments}
\label{subsec:Experiments}

In this section, we present high-level results of the our simple baselines, and our full method PANDA-EWC, (PANDA-SES can be found in Sec.\ref{sec:DetailedResults}) compared to the state-of-the-art: One-class SVM \cite{scholkopf2000support}, DeepSVDD \cite{ruff2018deep}, Multi-Head RotNet \cite{hendrycks2019using}. All the results of others that were available in the original papers were copied exactly. In cases that the result was not available, we run the experiments ourselves (where possible). As Joint Optimization requires extra data, we did not add it to this table, but compare and outperform it in Tab.~\ref{adapt_comp}. We compare our PANDA-OE to the OE baseline in \cite{hendrycks2019using} on CIFAR10, as the code or results for other classes were unavailable. Note that unless specifically mentioned otherwise, PANDA results were run with kNN. PANDA-OE used the original classifier - which performs a little better than kNN. We compare SPADE and relevant state-of-the-art baselines on anomaly segmentation.

We evaluated our method on a wide range of datasets (Tab.~\ref{datasets} Fig.~\ref{fig:dataset_images}) demonstrating different challenges in image anomaly detection (see App.~\ref{sm:dataset}).

\subsection{High-Level Results}
\label{subsec:High_Image}

\begin{table}[t]
  \caption{Details of datasets used for evaluation - number of classes, and average number of normal train and (normal and anomalous) test images per-class}
  \label{datasets}
  \centering
  \begin{tabular}{lllll}
    \toprule
Dataset	&	$N_{classes}$ 	&	$N_{train}$	&	$N_{test}$  \\
    \cmidrule(r){1-1}
    \cmidrule(r){2-4}
CIFAR10	&	10	&	5,000	&	10,000	\\
Fashion MNIST	&	10	&	6,000	&	10,000	\\
CIFAR100	&	20	&	2,500	&	10,000	\\
Flowers	&	102	&	10	&	7,169	\\
Birds	&	200	&	30	&	5,794	\\
CatsVsDogs	&	2	&	10,000	&	5,000	\\
MVTec	&	15	&	242	&	1,725	\\
WBC	&	4	&	59	&	62	\\
DIOR	&	19	&	649	&	9,243	\\
    \bottomrule
  \end{tabular}
  \vspace{-1.0em}
\end{table}

The main results are i) pretrained features achieve significantly better results than self-supervised features on all datasets, both in anomaly detection and segmentation. ii) Feature adaptation significantly improves the performance on larger datasets iii) Outlier exposure can further improve performance in the case where the given outliers are more similar to the anomalies than the normal data. OE achieves near perfect performance on CIFAR10/100 but hurts performance for Fashion MNIST/CatsVsDogs which are less similar to the 80M Tiny images dataset. 

A detailed analysis of the reason for better performance for each of these methods and an examination of its appropriateness will be presented in Sec.~\ref{sec:DetailedResults}.

\begin{table*}[]
  \caption{Anomaly detection performance  (Average ROC AUC \%)}
  \label{tab:big_dataset}
  \centering
  \begin{tabular}{lccccccc}
    \toprule
Dataset	& \multicolumn{3}{c}{Self-Supervised} 	&	\multicolumn{2}{c}{Pretrained} &	\multicolumn{2}{c}{OE}	\\
\cmidrule(r){1-1}
\cmidrule(r){2-4}
\cmidrule(r){5-6}
\cmidrule(r){7-8}
	&	OC-SVM	&	DeepSVDD	&	MHRot	&	 DN2 & PANDA &	MHRot &	PANDA-OE	\\									\cmidrule(r){1-1}
    \cmidrule(r){2-6}
    \cmidrule(r){7-8}    
CIFAR10
	&	64.7	&	64.8	&	90.1	&	92.5	&	\textbf{96.2}	&  95.6 &	\textbf{98.9}	\\
CIFAR100
	&	62.6	&	67.0	&	80.1	&	\textbf{94.1}	&	\textbf{94.1}		&  - & \textbf{97.3}	\\
FMNIST	&	92.8	&	84.8	&	93.2	&	94.5	&	\textbf{95.6}	& -  &	91.8	\\
CatsVsDogs	&	51.7	&	50.5	&	86.0	&	96.0	&	\textbf{97.3}	&  - & 94.5	 \\
DIOR& 70.7 & 70.0		&	73.3	&	93.0	&	\textbf{94.3} & - & \textbf{95.9} \\
    \bottomrule
  \end{tabular}
\vspace{-1.0em}
\end{table*}

\subsection{Analysis and Further Evaluation} \label{sec:DetailedResults}


\subsubsection{An analysis of feature representations}
\label{subsec:anl_representation} 

\textbf{A comparison of self-supervised and pretrained features}: In Tab.~\ref{tab:big_dataset} and Tab.~\ref{different_dataset}, we present a comparison between methods that use self-supervised and pretrained feature representations. 
We see that the autoencoder used by DeepSVDD is particularly poor. The results of the MHRotNet as a feature extractor are better, but still underperform PANDA methods (see App.~\ref{AppA} for more details). The performance of the raw deep ResNet features without adaptation significantly outperforms all methods, including Fashion MNIST and DIOR which have significant differences from the ImageNet dataset. We can therefore conclude that ImageNet-pretrained features typically have significant advantages over self-supervised features. Tab.~\ref{different_dataset} shows that self-supervised methods do not perform well on small datasets as such methods require large numbers of normal samples in order to learn strong features. On the other hand ImageNet-pretrained features obtain very strong results.

\textbf{Does the superiority of pretrained features extend to very different domains?}
The results in Tab.~\ref{different_dataset} on FMNIST, DIOR, WBC, MVTec suggest that out-of-domain pretrained features are better at anomaly detection than in-domain self-supervised features. We tested on datasets of various sizes, domains, resolutions and symmetries. On all those datasets pretrained features outperformed the SOTA. These datasets include significantly different objects from those of ImageNet, but also fine-grained intra-object anomalies, and represent a spectrum of data types: aerial images, microscopy, industrial images. This shows that one of the main arguments against using pretrained features, generalizing to distant domains, is not an issue in practice.

\textbf{Our simple pretrained feature baseline (SPADE) is extremely effective for anomaly segmentation:}
In Tab.\ref{tab:spade_comp} we can see that our simple, no-training, baseline (named SPADE) outperforms previous methods for anomaly segmentation, including those that use trained and pretrained features (see App.~\ref{app:implementation},\ref{app:spade} for metrics, specifications and detailed results). While we suspect feature adaptation can be used for further performance gain even for anomaly segmentation, we find that the MVTec dataset is too small for significant feature adaptation using the compactness loss. We believe that feature adaptation for segmentation calls for new adaptation methods, this is left for future work.

\begin{table}
  \caption{Pretrained feature performance on various small datasets  (Average ROC AUC \%)}
  \label{different_dataset}
  \centering
  \begin{tabular}{lccccc}
    \toprule
Dataset	& \multicolumn{3}{c}{Self-Supervised} 	&	\multicolumn{2}{c}{ Pretrained} 	\\
\cmidrule(r){1-1}
\cmidrule(r){2-4}
\cmidrule(r){5-5}
	&	OCSVM	&	DeepSVDD	&	MHRot	&	DN2  		\\									\cmidrule(r){1-1}
    \cmidrule(r){2-5}
Birds	&	62.0	&	60.8	&	64.4	&	\textbf{95.3} 		\\
Flowers &	74.5	&	78.1	&	65.9	&	\textbf{94.1} 	\\
MvTec	&	70.8	&	77.9	&	65.5 &	\textbf{86.5} 	 	\\
WBC	&	75.4	&	71.2	&	57.7	&	\textbf{87.4} 	\\

    \bottomrule
  \end{tabular}
  \vspace{-1.5em}
\end{table}

\begin{table*}[t]
  \caption{  Comparison of anomaly segmentation methods (pixel-level ROCAUC and PRO \%)}
  \label{tab:spade_comp}
  \centering
  \begin{tabular}{lllllllllllll}
    \toprule
	&	$AE_{SSIM}$ \cite{bergmann2019mvtec}	&	$AE_{L2}$ \cite{bergmann2019mvtec}	&	AnoGAN \cite{schlegl2017unsupervised}	&	CNN Dict \cite{napoletano2018anomaly}	&	CAVGA-$R_u$ \cite{venkataramanan2020attention}	&	Student \cite{bergmann2020uninformed}	&	SPADE	\\
    \cmidrule(r){1-1}
    \cmidrule(r){2-12}

ROCAUC	&	87	&	82	&	74	&	78	&	89	&	-	&	 \textbf{96.2}	\\
PRO	&	69.4	&	79	&	-	&	51.5	&	-	&	85.7	&	\textbf{92.1}	\\
	
    \bottomrule
  \end{tabular}
\end{table*}

\textbf{On the different supervision settings for one-class anomaly detection:} Anomaly detection methods employ different levels of supervision. Within the one-class classification task, one may use outlier exposure (OE) - an external dataset (e.g. ImageNet), pretrained features, or no external supervision at all. The most extensive supervision is used by OE, which requires a large external dataset at training time, and performs well only when such a dataset is from a similar domain to the anomalies (see Tab.~\ref{tab:big_dataset}). In cases where the dataset used for OE has significantly different properties, the network may not learn to distinguish between normal and anomalous data, as the normal and anomalous data may have more in common than the OE dataset. E.g. both normal and anomalous classes of Fashion MNIST are grayscale, OE using 80M Tiny Images will not be helpful, as the network may learn to classify only according to color. Pretrained features further improve OE, in cases where is suitable e.g. CIFAR10.

Pretraining, like Outlier Exposure, is also achieved through an external labelled dataset, but differently from OE, the external dataset is only required once - at the pretraining stage and is not used again. Additionally, the same features are applicable for very different image domains from that of the pretraining dataset (e.g. Fashion MNIST - grayscale images, DIOR - aerial images, WBC- medical images, MVTec - industrial images). Self supervised feature learning requires no external dataset at all, which can potentially be an advantage. While there might be image anomaly detection tasks where ImageNet-pretrained weights are not applicable, we saw no evidence for such cases after examining a broad spectrum of domains and datasets (Tab.~\ref{datasets}). This indicates that the extra supervision of the ImageNet-pretrained weights comes at virtually no cost.


\textbf{Can pretrained features boost the performance of RotNet-based methods?} We did not find evidence that pretrained features improve the performance of RotNet-based AD methods such as \cite{hendrycks2019using} (CIFAR10: 90.1\% vs. 86.6\% without and with pretraining). As can be seen in Tab.~\ref{trans_acc}, pretrained features improve the auxiliary task performance on the normal data, but also on the anomalous samples. As such methods rely on a generalization gap between normal and anomalous samples, deep features actually reduce this gap, as a solution to the auxiliary task becomes feasible for both types of images. For a more detailed analysis see App.~\ref{AppA}. 



\begin{table}
  \caption{Comparison of average transformation prediction accuracy  (\%), horiz. = horizontal, rot. = rotation.}
  \label{trans_acc}
  \centering
  \begin{tabular}{ccccc}
    \toprule
Method	&	\multicolumn{2}{c}{Normal}		&	\multicolumn{2}{c}{Anomalous} \\
\cmidrule(lr){1-1}  \cmidrule(lr){2-3}  \cmidrule(lr){4-5}
	&	Horiz.	&	Rot.	&	Horiz.	&	Rot.	\\
	\cmidrule(r){2-3}
	\cmidrule(r){4-5}  
Self-supervised	&	94.0	&	94.0	&	67.9	&	51.6	\\
Pretrained	&	94.4	&	92.3	&	71.4	&	61.3	\\
    \bottomrule
  \end{tabular}
\end{table}

			
	

\subsubsection{Feature adaptation methods}\label{adapt_anl}

\textbf{Benefits of feature adaptation:} Feature adaptation aims to make the distribution of the normal samples more compact, w.r.t. the anomalous samples. Our approach of finetuning pretrained features for compactness under EWC regularization, significantly improves the performance over "raw" pretrained features (see Tab.\ref{tab:big_dataset}). While the distance from the normal train samples' center, of both normal and anomalous test samples is reduced (see Fig.\ref{time_plots}), the average distance from the center of anomalous test samples is typically higher than that of normal samples, in relative terms, which makes anomalies easier to detect.

While PANDA-EWC may train more than $7.8k$ mini-batches without catastrophic collapse on CIFAR10, performance of training without regularization usually peaks higher but collapse earlier. We therefore set our constant early stopping epoch such that the net trains with to $2.3k$ minibatches on all datasets for comparison. Our PANDA-SES method usually achieves an anomaly score not far from the unregularized early stopping peak performance, but is most important in cases where unregularized training fails.


\textbf{A comparison of feature adaptation methods:}
In Tab.~\ref{adapt_comp} we compare PANDA against  (i) JO \cite{perera2019learning} - co-training compactness with ImageNet classification which requires ImageNet data at training time. We can see that PANDA - EWC always outperforms JO feature adaptation.  (ii) PANDA early stopping, generally has higher performance than PANDA-EWC, but has severe collapse issues on some classes. (iii) PANDA-SES is similar to early stopping, but PANDA-SES does not collapse as badly on CatsVsDogs dataset. We note that replacing the Fisher matrix by  equally weighting the changes in all parameters ( $\sum_i( \theta_i - \theta_i^*)  ^2$ ) achieves similar results to early stopping.

\begin{table}
  \caption{  A comparison of different feature adaptation methods (Avg. ROC AUC \%)}
  \label{adapt_comp}
  \centering
  \begin{tabular}{lcccc}
    \toprule
Dataset	&	Baseline	&	& PANDA			\\
    \cmidrule(r){1-1}
    \cmidrule(r){2-2}
    \cmidrule(r){3-5}
	&	JO	&	Early  & SES  &	 EWC		\\
    \midrule

CIFAR10 & 93.2 & \textbf{96.2} & 95.9 & \textbf{96.2}  \\
CIFAR100 & 91.1 & \textbf{94.8} & 94.6 & 94.1 \\
FMNIST & 94.9 & 95.4 & 95.5 & \textbf{95.6}  \\
CatsVsDogs & 96.1 & 91.9 & 95.7 & \textbf{97.3}  \\
DIOR & 93.1 & 95.4 & \textbf{95.6} & 94.3 \\  
    
	
    \bottomrule
  \end{tabular}
\end{table}

\textbf{Which are the best layers to finetune?}
Fine-tuning all layers is prone to feature collapse, even with continual learning. We therefore recommend finetuning only layers 3 \& 4  (see ablation in App.~\ref{app:comparisons}).

\textbf{DeepSVDD architectural changes:} DeepSVDD \cite{ruff2018deep} proposes various architectural changes, such as removing the bias parameters from the network, to prevent collapse to trivial features. To understand whether DeepSVDD gains its significant performance from its pretrained features or from its feature adaptation, we tried to replace its feature adaptation by closed-form linear data whitening. For both pretrained features and anomaly scoring, we used the DeepSVDD original code \cite{ruff2018deep}. We found empirically that the results obtained by the constrained architecture were about the same as those achieved with simple whitening of the data ($64.8\%$ vs. $64.6\%$, see App.~\ref{app:comparisons}). We ablated DeepSVDD by running it with the original LeNet (including biases) and found this did not deteriorate its anomaly detection performance. As architectural modifications are not the focus of this work, further investigation into architectures less prone to feature collapse is left for future work.

\subsubsection{Anomaly scoring functions}
\label{sec:scoring}

\textbf{Does kNN improve over distance to the center?} kNN achieves an improvement of around ~2\% on average w.r.t. to distance to the center (CIFAR10: 94.2\% vs 96.2\%).

\textbf{Can we improve over the linear complexity of kNN?} A naive implementation of kNN has linear runtime complexity in the number of training samples. 
For anomaly segmentation, approximating all the training sample features by $50$ means a speeds the method from $2.7$ frames-per-second to $41$ frames-per-second (faster than real-time), with $\sim$0.5\% ROCAUC decrease. For anomaly detection, even for very large datasets, or many thousands of means, both kNN and K-means can run faster than real-time.


\section{Conclusion and Outlook}\label{conclusion}

We first proposed simple baseline methods for anomaly detection and segmentation, that outperform the state-of-the-art. We further improved over the strong baselines by proposing a method that adapts pretrained features and mitigates catastrophic collapse. We showed that our results significantly outperform current methods while addressing their limitations. We analysed the reasons for the strong performance of our method.  
We note that the question of the optimal performance on image anomaly detection without ever having access to auxiliary data is unaddressed here, however we believe it is of mostly pure academic interest.

The main limitation of this work is the requirement for strong pretrained feature extractors. Much work was done on transferable image and text features and it is likely that current extractors can be effective to obtain features for time series and audio data as well. Generic feature extractors are not currently available for all data modalities and their development is an exciting direction for future work.

\section*{Acknowledgements}

This work was partly supported by the Federmann Cyber Security Research Center in conjunction with the Israel National Cyber Directorate.

\putbib

\clearpage

\end{bibunit}

\begin{bibunit}

\appendix
\part{Appendix} 
\addcontentsline{toc}{section}{Appendix} 
\parttoc 

\section{Pretrained Features, RotNet Auxiliary Tasks and Generalization} \label{AppA}
Let us take a closer look at the application of RotNet-based methods for image anomaly detection. We will venture to understand why initializing RotNets with pretrained features may actually impair their anomaly detection performance. In such cases, a network for rotation classification is trained on normal samples, and used to classify the rotation (and translations) applied to a test image. Each test image is checked for its rotation prediction accuracy, which is assumed to be worse for an anomalous images than for a typical normal image.

To correctly classify a rotation of a new image, the network may use traits within the image that are associated with its correct alignment. Such features may be associated with the normal class, or with the entire dataset (common to both the anomalous classes together). For illustrative purposes, let us consider a normal class with images containing a deer, and the anomalous class with images containing a horse. The horns of the deer may indicate the "upward" direction, but so does the position of the sky in the image, which is often sufficient to classify the rotation correctly. As shown in Tab.5 (in the main text), when initialized with pretrained features, the RotNet network achieves very good performance on the auxiliary tasks, both within and outside the normal class, indicating the use the more general traits that are common to more classes. 

Although at first sight it may appear that the improved auxiliary task performance should improve the performance on anomaly detection, this is in fact not the case! The reason is that features that generalize better, achieve better performance on the auxiliary task for anomalous data. The gap between the performance of normal and anomalous images of the auxiliary tasks, will therefore be smaller than with randomly-initialized networks - leading to degraded anomaly detection performance. For example, consider the illustrative case described above. A RotNet network that "overfits" to work only on the normal class deer, relying on the horns of the deer would classify rotations more accurately on deer images than on horse images (as its main feature is horns). On the other hand, a RotNet that also uses more general traits will achieve higher accuracy for both deer and horse images, leading to smaller a performance gap and inferior anomaly detection.

The above argument can be formulated using mutual information: In cases where the additional traits which are unique to the class do not add much information regarding the correct rotation, over the general features common to many classes, the class will have limited mutual information with the predicted rotation (conditional on the information already given by traits common to the entire datasets). When the conditional mutual information between the predicted rotation and the class traits decreases, we expect the predicted rotation to be less discriminative for anomaly detection, as we indeed see in Tab.\ref{Pretrained_rot}.

It is interesting to note that using features learned with RotNet for our transfer learning approach achieves inferior results to both MHRot and our method. Only through an ensemble of all rotations, as MHRot does, it achieves strong performance comparable to the MHRot performance. MHRot achieved $89.7\%$ in our re-implementation. Using the MHRot features as $\psi_0$, we compute the  kNN distance of the unadapted features between the test set images and the train set image transformed by the same transformation. When ensembling the 36 transformations - and using the average kNN distance, yields $88.7\%$. Another metric we examined is computing the average kNN distance between test data transformed under a specific transformation and the training set transformed by another transformation. Using the average same-transformation kNN distance minus the average different transformation kNN distance, achieves $89.8\%$ - a little better than the RotNet performance.

\begin{table*}[t]
  \caption{  Pretrained vs. raw initialization anomaly detection performance   (ROC AUC \%)}
  \label{Pretrained_rot}
  \centering
  \begin{tabular}{llllllllllll}
    \toprule
CIFAR10 class	&	0	&	1	&	2	&	3	&	4	&	5	&	6	&	7	&	8	&	9	&	Avg	\\
    \cmidrule(r){1-1}
    \cmidrule(r){2-11}
    \cmidrule(r){12-12}
			
Pretrained MHRot   &   70.1	&	   93.7	&	  84.4	&	   76.1	&	   89.7	&	  87.3	&	   91.1	&	   94.4	&	  86.8	&	  90.8	&	   86.4			       \\
MHRot	&	77.5	&	   96.9	&	   87.3	&	   80.9	&	  92.7	&	   90.2	&	   90.9	&	   96.5	&	   95.2	&	   93.3	&	   90.1		    \\
	
    \bottomrule
  \end{tabular}
\end{table*}

\section{Implementation Details}\label{app:implementation}

\subsection{PANDA}

\textit{Optimization:} We finetune the two last blocks of an ImageNet pretrained ResNet152 using SGD optimizer with weight decay of $w = 5\cdot10^{-5}$, and momentum of $m = 0.9$. We use $G = 10^{-3}$ gradient clipping. To have a comparable amount of training in the different dataset. We define the duration of each of our train using a constant number of minibatches, $32$ samples each.

\textit{EWC:} We use the fisher information matrix as obtained by \cite{kirkpatrick2017overcoming}, as explained in Sec.~\ref{Framework} in the main text. We weight the EWC loss with $\lambda = 10^{4}$. After obtaining EWC regularization, we train our net training on $7.8k$ minibatches.

\textit{Early stopping/Sample-wise early stopping:} We save a copy of the net every $5$ epochs. For early stopping we used the copy trained on $2.3k$ minibatches. For sample-wise early stopping we try all copies trained on up to $150k$ image samples (including repetitions).

\textit{Anomaly scoring:} Unless specified otherwise, we score the anomalies according to the kNN method with $k=2$ nearest neighbours. 

\textit{SES distance normalization:} When comparing different networks as in PANDA-SES method, we normalize each set of features by the typical kNN distance of its normal train features. To obtain the typical normal distance we would like to compute the average on the normal samples. However, computing the distance between normal training data has an issue: each point is its own nearest neighbour. Instead, we split the train set features ($90\%$ vs. $10\%$), and compute the kNN between the $10\%$ validation images and the gallery $90\%$ images.

\textit{PANDA Outlier Exposure}: The method was described in Sec.3 of the main text. For synthetic outlier images, we used the first 48k images of 80 Million Tiny Images \cite{torralba200880} with CIFAR10 and CIFAR100 images removed. We finetune the last block of an ImageNet pretrained ResNet152 with SGD optimizer using $75$ epochs and the following parameters: \textit{learning rate}: $0.1$ with \textit{gradient clipping}: {1e-3}, \textit{momentum}: $0.9$, and no weight decay.

\subsection{Anomaly Detection Baselines}

 We compare to the following methods:
 
\textit{OC-SVM}: One-class SVM with the RBF kernel. The hyper-parameters ($\nu\in\{0.1,\ldots, 0.9\},\gamma\in\{2^{-7},\ldots,2^2\}$) were optimized to maximize ROCAUC. 

\textit{DeepSVDD}: We resize all the images to $32 \times 32$ pixels and use the official pyTorch implementation with the CIFAR10 configuration.

\textit{MHRot} \cite{hendrycks2019using}: An improved version of the original RotNet approach. For high-resolution images we used the current GitHub implementation. For low resolution images, we modified the code to the architecture described in the paper, replicating the numbers in the paper on CIFAR10.

\textit{Outlier Exposure (MHRot)}: We use the outlier exposure performance as reported in \cite{hendrycks2019using}.

\subsection{SPADE}

\textit{Architecture:} In all experiments, we use a Wide-ResNet$50\times2$ feature extractor, which was pretrained on ImageNet. 

\textit{Resolution:} MVTec images were resized to $256 \times 256$ and cropped to $224 \times 224$.  All metrics were calculated at $256 \times 256$ image resolution, and we used cv2.INTERAREA for resizing when needed. 

\textit{Layers:} Unless otherwise specified, we used features from the ResNet at the end of the first block ($56 \times 56$), second block ($28 \times 28$) and third block ($14 \times 14$), all with equal weights.   In Tab.~\ref{tab:ablation:level} we compare different level of the feature pyramid as feature descriptor. We experienced that using activations of too high resolution ($56 \times 56$) significantly hurts performance due to limited context, while using the higher levels on their own, results in diminished performance (due to lower resolution). Using a combination of all three upstream layers in the pyramid results in the best performance.

\textit{Combining features from different layers:} We evaluated two ways of combining per-pixel features extracted from different layers. Concatenation - resampling the activation to the same resolutions and concatenating all per-pixel features to form a combined feature. Ensembling - computing the per-pixel anomaly score using the per-pixel feature of each layer, and adding the per-pixel per-layer scores of all layers to form a combined score. We found the ensemble approach was more robust and yielded a bit better results. Therefore, we report it.

\textit{Postprocessing:} After computing the pixel-wise anomaly score for each image, we smoothed the results with a Gaussian filter ($\sigma = 5$).

For fast nearest neighbour implementation, we used the \textit{FAISS} library \cite{johnson2019billion}.

\begin{table*}[!h]
  \centering
  \caption{Anomaly segmentation accuracy on MVTec with different ResNet layers (PRO $\%$)}
  \label{tab:ablation:level}

    \begin{tabular}{lcccccc}
        \toprule

Summed layers	&	Layers 0,1,2	&	Layer 0	&	Layer 1	&	Layer 2	\\

\cmidrule(r){1-1}
\cmidrule(r){2-5}

Carpet	&	98.9	&	86.7	&	97.9	&	98.8	\\
Grid	&	97.6	&	98.9	&	98.9	&	96.3	\\
Leather	&	99.1	&	98.3	&	99.3	&	99.0	\\
Tile	&	94.4	&	80.5	&	91.1	&	94.2	\\
Wood	&	93.7	&	92.3	&	94.5	&	92.5	\\
Bottle	&	98.1	&	89.6	&	98.0	&	97.9	\\
Cable	&	96.4	&	70.8	&	93.5	&	96.9	\\
Capsule	&	99.0	&	97.3	&	98.8	&	98.7	\\
Hazelnut	&	98.6	&	96.6	&	97.6	&	98.6	\\
Metal nut	&	97.4	&	88.4	&	97.0	&	96.7	\\
Pill	&	96.4	&	95.9	&	95.8	&	95.9	\\
Screw	&	99.2	&	98.8	&	99.4	&	98.6	\\
Toothbrush	&	98.8	&	96.9	&	98.8	&	98.6	\\
Transistor	&	94.2	&	71.0	&	84.0	&	95.9	\\
Zipper	&	98.1	&	96.4	&	97.9	&	97.7	\\
									
\cmidrule(r){1-1}
\cmidrule(r){2-5}								
Average	&	\textbf{97.3}	&	90.6	&	96.2	&	97.1	\\
	 \bottomrule
    \end{tabular}
\end{table*}

\section{Datasets}
\label{sm:dataset}
\textbf{Standard datasets:} We evaluate our method on a set of commonly used datasets: \textit{CIFAR10} \cite{krizhevsky2009learning}:  Consists of RGB images of 10  object classes. \textit{Fashion MNIST} \cite{xiao2017fashion}:  Consists of grayscale images of 10 fashion item classes. \textit{CIFAR100} \cite{krizhevsky2009learning}:  We use the coarse-grained version that consists of 20 classes. \textit{DogsVsCats}: High resolution color images of two classes: cats and dogs. The data were extracted from the ASIRRA dataset\cite{elson2007asirra}, we split each class to the first 10,000 images as train and the last 2,500 as test.

\textbf{Small datasets:} We report results on several small datasets from different domains: \textit{$102$ Category Flowers \& Caltech-UCSD Birds $200$} \cite{nilsback2008automated} \cite{wah2011caltech}: For each of those datasets we evaluated the methods using only the first $20$ classes as normal train set, and using the entire test set for evaluation. \textit{MVTec} \cite{bergmann2019mvtec}: This datasets contain $15$ different industrial products, with normal images of proper products for train and $1-9$ types of manufacturing errors as anomalies. The anomalies in MVTec are in-class i.e. the anomalous images come from the same class of normal images with subtle variations. We also use the MVTec dataset for the anomaly segmentation results.

\textbf{Symmetric datasets:} We evaluated our method on datasets that contain symmetries, such as images that have no preferred angle (microscopy, aerial images.): \textit{WBC} \cite{zheng2018fast}: We used the $4$ big classes in \textit{"Dataset 1"} of microscopy images of white blood cells, and a $80\%/20\%$ train-test split. \textit{DIOR} \cite{li2020object}: We preprocessed the DIOR aerial image dataset by taking the segmented object in classes that have more than $50$ images with size larger than $120\times120$ pixels. We can see that RotNet-type methods perform particularly poorly on such datasets.

\section{Choosing the Layers to Finetune} \label{app:comparisons}

Fine-tuning all layers is prone to feature collapse, even with continual learning (see Tab.\ref{blocks_compare}). Finetuning Blocks 3 \& 4, or 2, 3 \& 4, results in similar performance. Finetuning only block 4 results in similar performance to linear whitening of the features according to the train samples (94.6 with whitening vs. 94.8 with finetuning only the last block). Similar effect as can be seen in the original DeepSVDD architecture (see Tab.\ref{svdd_comp}). We therefore recommend finetuning Blocks 3\&4.


\begin{table}[t]
  \caption{  Performance  of finetuning different ResNet blocks  (CIFAR10 w. EWC, ROC AUC \%)}
  \label{blocks_compare}
  \centering
  \begin{tabular}{llllllllllll}
    \toprule
    \multicolumn{7}{c}{with std}                   \\
    \cmidrule(r){4-6}
Trained Blocks	&	1,2,3,4	&	2,3,4	&	3,4	&	4		\\
\cmidrule(r){1-1}
\cmidrule(r){2-5}
			
Avg	& 94.9  &  	95.9  &  	\textbf{96.2}  &  	94.8   \\

    \bottomrule
  \end{tabular}
  \vspace{-1.25em}
\end{table}

\begin{table*}[t]
  \caption{  Deep SVDD vs. PCA Whitening Anomaly Detection Performance   (ROC AUC \%)}
  \label{svdd_comp}
  \centering
  \begin{tabular}{llllllllllll}
    \toprule
    \multicolumn{7}{c}{with std}                   \\
    \cmidrule(r){4-6}
CIFAR10 class	&	0	&	1	&	2	&	3	&	4	&	5	&	6	&	7	&	8	&	9	&	Avg	\\
    \cmidrule(r){1-1}
    \cmidrule(r){2-11}
    \cmidrule(r){12-12}
			
PCA whitening	& 62.0  &  	63.6  &  	49.7  &  	59.9  &  	59.8  &  	65.8  &  	68.3  &  	68.0  &  	75.5  &  	71.2  & 64.8 \\
Deep SVDD	& 59.7  &  	64.3  &  	48.4  &  	61.5  &  	61.3  &  	65.5  &  	70.1  &  	68.9  &  	75.3  &  	72.5  & 64.6 \\
	
    \bottomrule
  \end{tabular}
\end{table*}

\section{SPADE: Detailed Results}
\label{app:spade}
\begin{figure*}[t]
\begin{center}
    \begin{tabular}{cccc}
     \includegraphics[scale=0.33]{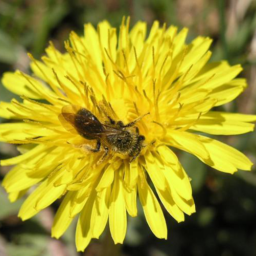} &
   \includegraphics[scale=0.376]{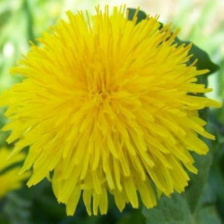} &
   \includegraphics[scale=0.33]{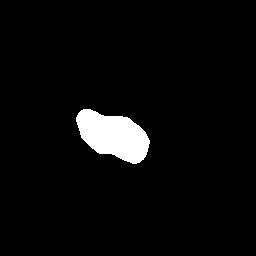} &
   \includegraphics[scale=0.33]{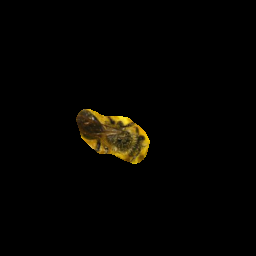} \\
        \includegraphics[scale=0.33]{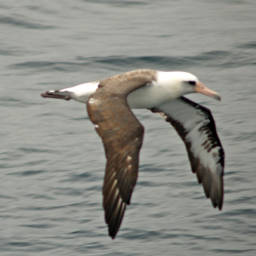} &
   \includegraphics[scale=0.376]{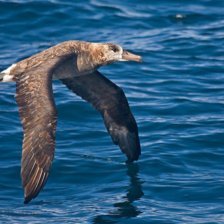} &
   \includegraphics[scale=0.33]{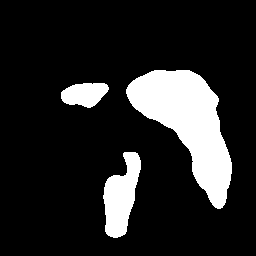} &
   \includegraphics[scale=0.33]{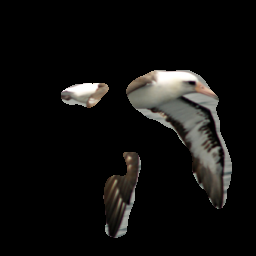} \\
       \includegraphics[scale=0.33]{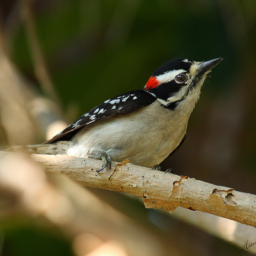} &
   \includegraphics[scale=0.376]{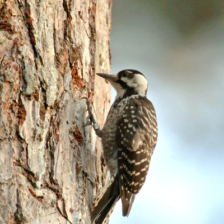} &
   \includegraphics[scale=0.33]{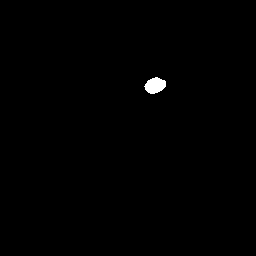} &
   \includegraphics[scale=0.33]{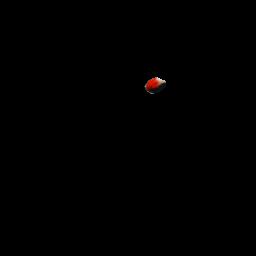}
   
    \end{tabular}
    \end{center}
    \caption{ An evaluation of SPADE on detecting anomalies between flowers with or without insects (taken from one category of 102 Category Flower Dataset \cite{nilsback2008automated}) and bird varieties (taken from Caltech-UCSD Birds 200) \cite{welinder2010caltech}.   (left to right) i) An anomalous image ii) A normal train set image iii) The mask detected by SPADE iv) The predicted anomalous image pixels. SPADE was able to detect the insect on the anomalous flower (top), the white colors of the anomalous albatross (center) and the red spot on the anomalous bird (bottom). }
    \label{fig:flower}
\end{figure*}

In this section, we report the full results for SPADE and its relevant baselines. We evaluate our method using two established metrics. The first is per-pixel ROCAUC. The ROC curve is calculated by first computing the anomaly score of each pixel and then scanning over the range of thresholds, on pixels from all test images together. The anomalous category is designated as positive. It was noted by several previous works that ROCAUC is biased in favor of large anomalies. In order to reduce this bias, Bergmann et al \cite{bergmann2020uninformed} propose the PRO (per-region overlap) curve metric. They first separate anomaly masks into their connected components, therefore dividing them into individual anomaly regions. By changing the detection threshold, they scan over false positive rates (FPR), for each FPR they compute PRO i.e. the proportion of the pixels of each region that are detected as anomalous. The PRO score at this FPR is the average coverage across all anomalous regions. The PRO curve metric computes the integral across FPR rates from $0$ to $0.3$. The PRO score is the normalized value of this integral. We can see from Tab.~\ref{tab:mvtec_pixel_pro} and Tab.~\ref{tab:mvtec_pixel_roc} that our method significantly outperforms the baselines in terms of both metrics. Qualitative results of our method are presented in Fig.~\ref{fig:flower}.

\begin{table*}[t]
  \centering
  \caption{Sub-Image anomaly detection accuracy on MVTec (ROCAUC $\%$)}
  \label{tab:mvtec_pixel_roc}

    \begin{tabular}{lcccccccc}
    \toprule      

   & $AE_{SSIM}$  &  $AE_{L2}$  & AnoGAN  & CNN Dict & TI & VM & CAVGA-$R_u$  & SPADE\\
    \midrule
Carpet &	87	&	59	&	54	&	72	&	88	&	-	& -	&	98.6	\\
Grid &	94	&	90	&	58	&	59	&	72	&	-	& -	&	99.0	\\
Leather &	78	&	75	&	64	&	87	&	97	&	-	& -	&	99.5	\\
Tile &	59	&	51	&	50	&	93	&	41	&	-	& -	&	89.8	\\
Wood &	73	&	73	&	62	&	91	&	78	&	-	& -	&	95.8	\\
Bottle &	93	&	86	&	86	&	78	&	-	&	82	& -	&	98.1	\\
Cable &	82	&	86	&	78	&	79	&	-	&	-	& -	&	93.2	\\
Capsule &	94	&	88	&	84	&	84	&	-	&	76	& -	&	98.6	\\
Hazelnut &	97	&	95	&	87	&	72	&	-	&	-	& -	&	98.9	\\
Metal nut &	89	&	86	&	76	&	82	&	-	&	60	& -	&	96.9	\\
Pill &	91	&	85	&	87	&	68	&	-	&	83	& -	&	96.5	\\
Screw &	96	&	96	&	80	&	87	&	-	&	94	& -	&	99.5	\\
Toothbrush &	92	&	93	&	90	&	77	&		&	68	& -	&	98.9	\\
Transistor &	90	&	86	&	80	&	66	&	-	&	-	& -	&	81.0	\\
Zipper &	88	&	77	&	78	&	76	&	-	&	-	& -	&	98.8	\\
\midrule														
Average & 87	&	82	&	74	&	78	&	75	&	77	& 89	& \textbf{96.2}	\\
	 \bottomrule
    \end{tabular}
\end{table*}

\begin{table*}[t]
  \centering
  \caption{Sub-Image anomaly detection accuracy on MVTec (PRO $\%$)}
  \label{tab:mvtec_pixel_pro}

    \begin{tabular}{lcccccccc}
    \toprule      

    & Student & 1-NN & OC-SVM  & $\ell_2$-AE & VAE & SSIM-AE  & CNN-Dict & SPADE \\
    \midrule
Carpet	&			69.5	&	51.2	&	35.5	&	45.6	&	50.1	&	64.7	&	46.9	&	96.1		\\
Grid	&			81.9	&	22.8	&	12.5	&	58.2	&	22.4	&	84.9	&	18.3	&	97.0		\\
Leather	&			81.9	&	44.6	&	30.6	&	81.9	&	63.5	&	56.1	&	64.1	&	98.8		\\
Tile	&			91.2	&	82.2	&	72.2	&	89.7	&	87.0	&	17.5	&	79.7	&	77.1		\\
Wood	&			72.5	&	50.2	&	33.6	&	72.7	&	62.8	&	60.5	&	62.1	&	93.8		\\
Bottle	&			91.8	&	89.8	&	85.0	&	91.0	&	89.7	&	83.4	&	74.2	&	95.6		\\
Cable	&			86.5	&	80.6	&	43.1	&	82.5	&	65.4	&	47.8	&	55.8	&	85.3		\\
Capsule	&			91.6	&	63.1	&	55.4	&	86.2	&	52.6	&	86.0	&	30.6	&	95.5		\\
Hazelnut	&			93.7	&	86.1	&	61.6	&	91.7	&	87.8	&	91.6	&	84.4	&	94.8		\\
Metal nut	&			89.5	&	70.5	&	31.9	&	83.0	&	57.6	&	60.3	&	35.8	&	94.1		\\
Pill	&			93.5	&	72.5	&	54.4	&	89.3	&	76.9	&	83.0	&	46.0	&	96.2		\\
Screw	&			92.8	&	60.4	&	64.4	&	75.4	&	55.9	&	88.7	&	27.7	&	97.4		\\
Toothbrush	&			86.3	&	67.5	&	53.8	&	82.2	&	69.3	&	78.4	&	15.1	&	94.4		\\
Transistor	&			70.1	&	68.0	&	49.6	&	72.8	&	62.6	&	72.5	&	62.8	&	67.8		\\
Zipper	&			93.3	&	51.2	&	35.5	&	83.9	&	54.9	&	66.5	&	70.3	&	96.9		\\
\midrule																				
Average	&			85.7	&	64	&	47.9	&	79	&	63.9	&	69.4	&	51.5	&	\textbf{92.1}		\\																
	 \bottomrule
    \end{tabular}
\end{table*}

\clearpage

\putbib
\end{bibunit}


\begin{thebibliography}{10}\itemsep=-1pt

\bibitem{bergman2020classification}
Liron Bergman and Yedid Hoshen.
\newblock Classification-based anomaly detection for general data.
\newblock In {\em ICLR}, 2020.

\bibitem{bergmann2019mvtec}
Paul Bergmann, Michael Fauser, David Sattlegger, and Carsten Steger.
\newblock Mvtec ad--a comprehensive real-world dataset for unsupervised anomaly
  detection.
\newblock In {\em Proceedings of the IEEE Conference on Computer Vision and
  Pattern Recognition}, pages 9592--9600, 2019.

\bibitem{bergmann2020uninformed}
Paul Bergmann, Michael Fauser, David Sattlegger, and Carsten Steger.
\newblock Uninformed students: Student-teacher anomaly detection with
  discriminative latent embeddings.
\newblock In {\em Proceedings of the IEEE/CVF Conference on Computer Vision and
  Pattern Recognition}, pages 4183--4192, 2020.

\bibitem{candes2011robust}
Emmanuel~J Cand{\`e}s, Xiaodong Li, Yi Ma, and John Wright.
\newblock Robust principal component analysis?
\newblock {\em JACM}, 2011.

\bibitem{chen2020simple}
Ting Chen, Simon Kornblith, Mohammad Norouzi, and Geoffrey Hinton.
\newblock A simple framework for contrastive learning of visual
  representations.
\newblock {\em arXiv preprint arXiv:2002.05709}, 2020.

\bibitem{deecke2020deep}
Lucas Deecke, Lukas Ruff, Robert~A Vandermeulen, and Hakan Bilen.
\newblock Deep anomaly detection by residual adaptation.
\newblock {\em arXiv preprint arXiv:2010.02310}, 2020.

\bibitem{dgroup}
Pierluca D’Oro, Ennio Nasca, Jonathan Masci, and Matteo Matteucci.
\newblock Group anomaly detection via graph autoencoders.
\newblock {\em NIPS Workshop}, 2019.

\bibitem{eskin2002geometric}
Eleazar Eskin, Andrew Arnold, Michael Prerau, Leonid Portnoy, and Sal Stolfo.
\newblock A geometric framework for unsupervised anomaly detection.
\newblock In {\em Applications of data mining in computer security}, pages
  77--101. Springer, 2002.

\bibitem{georgescu2021anomaly}
Mariana-Iuliana Georgescu, Antonio Barbalau, Radu~Tudor Ionescu, Fahad~Shahbaz
  Khan, Marius Popescu, and Mubarak Shah.
\newblock Anomaly detection in video via self-supervised and multi-task
  learning.
\newblock In {\em Proceedings of the IEEE/CVF Conference on Computer Vision and
  Pattern Recognition}, pages 12742--12752, 2021.

\bibitem{gidaris2018unsupervised}
Spyros Gidaris, Praveer Singh, and Nikos Komodakis.
\newblock Unsupervised representation learning by predicting image rotations.
\newblock {\em arXiv preprint arXiv:1803.07728}, 2018.

\bibitem{glodek2013ensemble}
Michael Glodek, Martin Schels, and Friedhelm Schwenker.
\newblock Ensemble gaussian mixture models for probability density estimation.
\newblock {\em Computational Statistics}, 28(1):127--138, 2013.

\bibitem{golan2018deep}
Izhak Golan and Ran El-Yaniv.
\newblock Deep anomaly detection using geometric transformations.
\newblock In {\em NeurIPS}, 2018.

\bibitem{goodfellow2014generative}
Ian~J Goodfellow, Jean Pouget-Abadie, Mehdi Mirza, Bing Xu, David Warde-Farley,
  Sherjil Ozair, Aaron Courville, and Yoshua Bengio.
\newblock Generative adversarial networks.
\newblock {\em arXiv preprint arXiv:1406.2661}, 2014.

\bibitem{hartigan1979algorithm}
John~A Hartigan and Manchek~A Wong.
\newblock Algorithm as 136: A k-means clustering algorithm.
\newblock {\em Journal of the Royal Statistical Society. Series C (Applied
  Statistics)}, 1979.

\bibitem{hendrycks2019pretraining}
Dan Hendrycks, Kimin Lee, and Mantas Mazeika.
\newblock Using pre-training can improve model robustness and uncertainty.
\newblock {\em arXiv preprint arXiv:1901.09960}, 2019.

\bibitem{hendrycks2018deep}
Dan Hendrycks, Mantas Mazeika, and Thomas Dietterich.
\newblock Deep anomaly detection with outlier exposure.
\newblock {\em arXiv preprint arXiv:1812.04606}, 2018.

\bibitem{hendrycks2019using}
Dan Hendrycks, Mantas Mazeika, Saurav Kadavath, and Dawn Song.
\newblock Using self-supervised learning can improve model robustness and
  uncertainty.
\newblock In {\em NeurIPS}, 2019.

\bibitem{huh2016makes}
Minyoung Huh, Pulkit Agrawal, and Alexei~A Efros.
\newblock What makes imagenet good for transfer learning?
\newblock {\em arXiv preprint arXiv:1608.08614}, 2016.

\bibitem{jolliffe2011principal}
Ian Jolliffe.
\newblock {\em Principal component analysis}.
\newblock Springer, 2011.

\bibitem{kirkpatrick2017overcoming}
James Kirkpatrick, Razvan Pascanu, Neil Rabinowitz, Joel Veness, Guillaume
  Desjardins, Andrei~A Rusu, Kieran Milan, John Quan, Tiago Ramalho, Agnieszka
  Grabska-Barwinska, et~al.
\newblock Overcoming catastrophic forgetting in neural networks.
\newblock {\em Proceedings of the national academy of sciences},
  114(13):3521--3526, 2017.

\bibitem{latecki2007outlier}
Longin~Jan Latecki, Aleksandar Lazarevic, and Dragoljub Pokrajac.
\newblock Outlier detection with kernel density functions.
\newblock In {\em International Workshop on Machine Learning and Data Mining in
  Pattern Recognition}, pages 61--75. Springer, 2007.

\bibitem{napoletano2018anomaly}
Paolo Napoletano, Flavio Piccoli, and Raimondo Schettini.
\newblock Anomaly detection in nanofibrous materials by cnn-based
  self-similarity.
\newblock {\em Sensors}, 18(1):209, 2018.

\bibitem{pang2021deep}
Guansong Pang, Chunhua Shen, Longbing Cao, and Anton Van~Den Hengel.
\newblock Deep learning for anomaly detection: A review.
\newblock {\em ACM Computing Surveys (CSUR)}, 54(2):1--38, 2021.

\bibitem{perera2019learning}
Pramuditha Perera and Vishal~M Patel.
\newblock Learning deep features for one-class classification.
\newblock {\em IEEE Transactions on Image Processing}, 28(11):5450--5463, 2019.

\bibitem{rippel2020modeling}
Oliver Rippel, Patrick Mertens, and Dorit Merhof.
\newblock Modeling the distribution of normal data in pre-trained deep features
  for anomaly detection.
\newblock {\em arXiv preprint arXiv:2005.14140}, 2020.

\bibitem{ruff2018deep}
Lukas Ruff, Nico Gornitz, Lucas Deecke, Shoaib~Ahmed Siddiqui, Robert
  Vandermeulen, Alexander Binder, Emmanuel M{\"u}ller, and Marius Kloft.
\newblock Deep one-class classification.
\newblock In {\em ICML}, 2018.

\bibitem{ruff2021unifying}
Lukas Ruff, Jacob~R Kauffmann, Robert~A Vandermeulen, Gr{\'e}goire Montavon,
  Wojciech Samek, Marius Kloft, Thomas~G Dietterich, and Klaus-Robert
  M{\"u}ller.
\newblock A unifying review of deep and shallow anomaly detection.
\newblock {\em Proceedings of the IEEE}, 2021.

\bibitem{schlegl2017unsupervised}
Thomas Schlegl, Philipp Seeb{\"o}ck, Sebastian~M Waldstein, Ursula
  Schmidt-Erfurth, and Georg Langs.
\newblock Unsupervised anomaly detection with generative adversarial networks
  to guide marker discovery.
\newblock In {\em International conference on information processing in medical
  imaging}, pages 146--157. Springer, 2017.

\bibitem{scholkopf2000support}
Bernhard Scholkopf, Robert~C Williamson, Alex~J Smola, John Shawe-Taylor, and
  John~C Platt.
\newblock Support vector method for novelty detection.
\newblock In {\em NIPS}, 2000.

\bibitem{selvaraju2017grad}
Ramprasaath~R Selvaraju, Michael Cogswell, Abhishek Das, Ramakrishna Vedantam,
  Devi Parikh, and Dhruv Batra.
\newblock Grad-cam: Visual explanations from deep networks via gradient-based
  localization.
\newblock In {\em Proceedings of the IEEE international conference on computer
  vision}, pages 618--626, 2017.

\bibitem{smeureanu2017deep}
Sorina Smeureanu, Radu~Tudor Ionescu, Marius Popescu, and Bogdan Alexe.
\newblock Deep appearance features for abnormal behavior detection in video.
\newblock In {\em International Conference on Image Analysis and Processing},
  pages 779--789. Springer, 2017.

\bibitem{venkataramanan2020attention}
Shashanka Venkataramanan, Kuan-Chuan Peng, Rajat~Vikram Singh, and Abhijit
  Mahalanobis.
\newblock Attention guided anomaly localization in images.
\newblock In {\em European Conference on Computer Vision}, pages 485--503.
  Springer, 2020.

\bibitem{zong2018deep}
Bo Zong, Qi Song, Martin~Renqiang Min, Wei Cheng, Cristian Lumezanu, Daeki Cho,
  and Haifeng Chen.
\newblock Deep autoencoding gaussian mixture model for unsupervised anomaly
  detection.
\newblock {\em ICLR}, 2018.

\end{thebibliography}


\begin{thebibliography}{10}\itemsep=-1pt

\bibitem{bergmann2019mvtec}
Paul Bergmann, Michael Fauser, David Sattlegger, and Carsten Steger.
\newblock Mvtec ad--a comprehensive real-world dataset for unsupervised anomaly
  detection.
\newblock In {\em Proceedings of the IEEE Conference on Computer Vision and
  Pattern Recognition}, pages 9592--9600, 2019.

\bibitem{bergmann2020uninformed}
Paul Bergmann, Michael Fauser, David Sattlegger, and Carsten Steger.
\newblock Uninformed students: Student-teacher anomaly detection with
  discriminative latent embeddings.
\newblock In {\em Proceedings of the IEEE/CVF Conference on Computer Vision and
  Pattern Recognition}, pages 4183--4192, 2020.

\bibitem{elson2007asirra}
Jeremy Elson, John~R Douceur, Jon Howell, and Jared Saul.
\newblock Asirra: a captcha that exploits interest-aligned manual image
  categorization.
\newblock In {\em ACM Conference on Computer and Communications Security},
  volume~7, pages 366--374, 2007.

\bibitem{hendrycks2019using}
Dan Hendrycks, Mantas Mazeika, Saurav Kadavath, and Dawn Song.
\newblock Using self-supervised learning can improve model robustness and
  uncertainty.
\newblock In {\em NeurIPS}, 2019.

\bibitem{johnson2019billion}
Jeff Johnson, Matthijs Douze, and Herv{\'e} J{\'e}gou.
\newblock Billion-scale similarity search with gpus.
\newblock {\em IEEE Transactions on Big Data}, 2019.

\bibitem{kirkpatrick2017overcoming}
James Kirkpatrick, Razvan Pascanu, Neil Rabinowitz, Joel Veness, Guillaume
  Desjardins, Andrei~A Rusu, Kieran Milan, John Quan, Tiago Ramalho, Agnieszka
  Grabska-Barwinska, et~al.
\newblock Overcoming catastrophic forgetting in neural networks.
\newblock {\em Proceedings of the national academy of sciences},
  114(13):3521--3526, 2017.

\bibitem{krizhevsky2009learning}
Alex Krizhevsky, Geoffrey Hinton, et~al.
\newblock Learning multiple layers of features from tiny images.
\newblock 2009.

\bibitem{li2020object}
Ke Li, Gang Wan, Gong Cheng, Liqiu Meng, and Junwei Han.
\newblock Object detection in optical remote sensing images: A survey and a new
  benchmark.
\newblock {\em ISPRS Journal of Photogrammetry and Remote Sensing},
  159:296--307, 2020.

\bibitem{nilsback2008automated}
Maria-Elena Nilsback and Andrew Zisserman.
\newblock Automated flower classification over a large number of classes.
\newblock In {\em 2008 Sixth Indian Conference on Computer Vision, Graphics \&
  Image Processing}, pages 722--729. IEEE, 2008.

\bibitem{torralba200880}
Antonio Torralba, Rob Fergus, and William~T Freeman.
\newblock 80 million tiny images: A large data set for nonparametric object and
  scene recognition.
\newblock {\em IEEE transactions on pattern analysis and machine intelligence},
  30(11):1958--1970, 2008.

\bibitem{wah2011caltech}
Catherine Wah, Steve Branson, Peter Welinder, Pietro Perona, and Serge
  Belongie.
\newblock The caltech-ucsd birds-200-2011 dataset.
\newblock 2011.

\bibitem{welinder2010caltech}
Peter Welinder, Steve Branson, Takeshi Mita, Catherine Wah, Florian Schroff,
  Serge Belongie, and Pietro Perona.
\newblock Caltech-ucsd birds 200.
\newblock 2010.

\bibitem{xiao2017fashion}
Han Xiao, Kashif Rasul, and Roland Vollgraf.
\newblock Fashion-mnist: a novel image dataset for benchmarking machine
  learning algorithms.
\newblock {\em arXiv preprint arXiv:1708.07747}, 2017.

\bibitem{zheng2018fast}
Xin Zheng, Yong Wang, Guoyou Wang, and Jianguo Liu.
\newblock Fast and robust segmentation of white blood cell images by
  self-supervised learning.
\newblock {\em Micron}, 107:55--71, 2018.

\end{thebibliography}
\end{document}